\documentclass[10pt,twocolumn,letterpaper]{article}

\usepackage[pagenumbers]{cvpr} 

\definecolor{cvprblue}{rgb}{0.21,0.49,0.74}
\usepackage[pagebackref,breaklinks,colorlinks,allcolors=cvprblue]{hyperref}
\usepackage{graphicx}
\usepackage{amsmath}
\usepackage{amssymb}
\usepackage{booktabs}
\usepackage{colortbl}
\usepackage{color}
\usepackage{enumitem}
\usepackage{mmstyles}
\usepackage{multirow}
\usepackage{makecell}
\usepackage{siunitx}
\usepackage{subcaption}
\usepackage{pifont}
\usepackage{xspace}
\usepackage[table]{xcolor}

\definecolor{ForestGreen}{rgb}{0.13, 0.55, 0.13}

\newcommand{\negval}[1]{\textcolor{red!70!black}{#1}}
\newcommand{\eqlval}[1]{\textcolor{gray!100!black}{#1}}
\newcommand{\xmark}{\ding{55}}

\graphicspath{{figures/}}
\DeclareGraphicsExtensions{.pdf}

\def\paperID{} 
\def\confName{CVPR}
\def\confYear{2026}

\title{SSL-R1: Self-Supervised Visual Reinforcement Post-Training for \\Multimodal Large Language Models}

\author{
Jiahao Xie$^{1,2}$, Alessio Tonioni$^{3}$, Nathalie Rauschmayr$^{3}$, Federico Tombari$^{3}$, Bernt Schiele$^{1,2}$\\
$^{1}$Max Planck Institute for Informatics, SIC \quad 
$^{2}$VIA Research Center \quad 
$^{3}$Google
}

\begin{document}

\twocolumn[{%
\renewcommand\twocolumn[1][]{#1}%
\maketitle
\begin{center}
    \centering 
    \includegraphics[width=\linewidth]{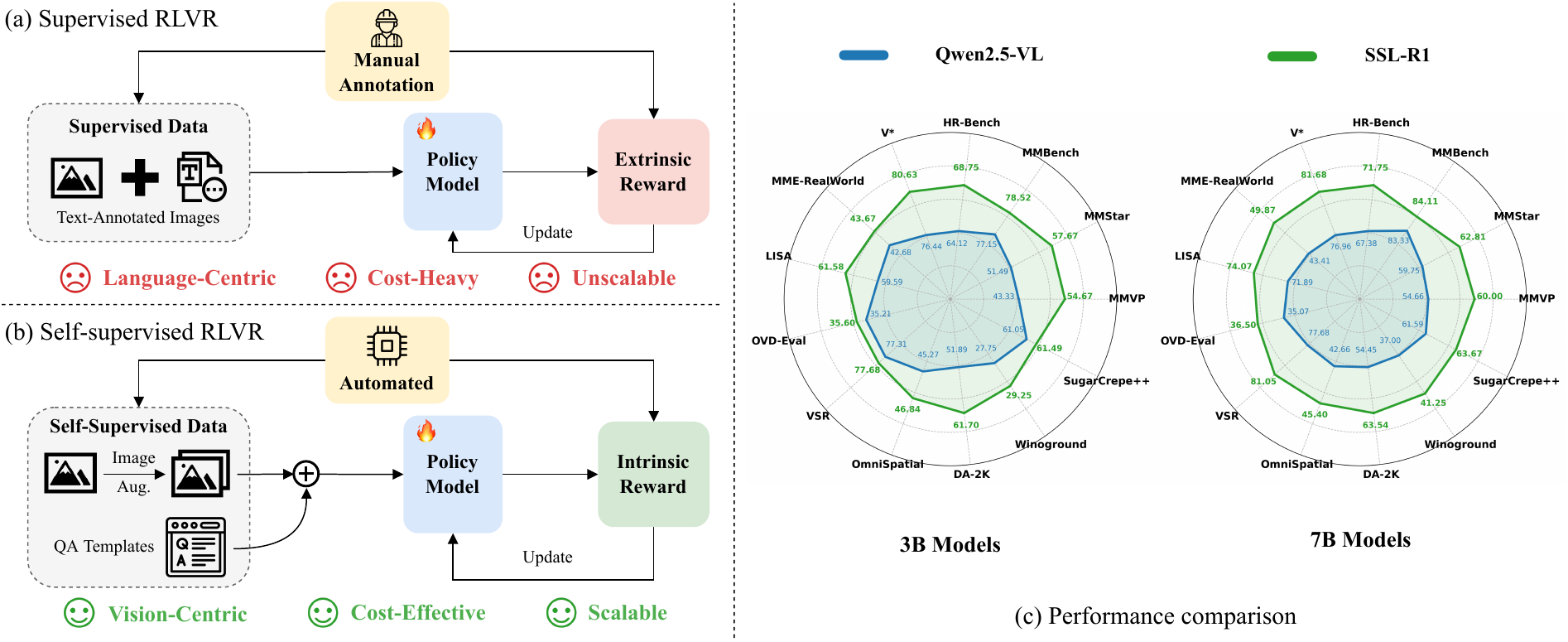}
    \captionof{figure}{
    \textbf{(a)} Existing reinforcement learning with verifiable rewards (RLVR) for post-training MLLMs are supervised, requiring a large volume of high-quality language-centric multimodal data with human annotations, which is very expensive and unsustainable. \textbf{(b)} We propose SSL-R1, a generic self-supervised RLVR-based post-training framework that derives intrinsically verifiable rewards from input images, requiring neither human nor external model supervision. \textbf{(c)} SSL-R1 is vision-centric, cost-effective, scalable, and consistently improves Qwen2.5-VL on 13 vision-centric multimodal benchmarks, spanning fine-grained perception, spatial understanding, and compositional understanding. Note that the value ranges differ across benchmarks in each radar plot for clarity.
    }
    \label{fig:teaser}
\end{center}%
}]


\begin{abstract}
Reinforcement learning (RL) with verifiable rewards (RLVR) has demonstrated the great potential of enhancing the reasoning abilities in multimodal large language models (MLLMs).
However, the reliance on language-centric priors and expensive manual annotations prevents MLLMs' intrinsic visual understanding and scalable reward designs.
In this work, we introduce SSL-R1, a generic self-supervised RL framework that derives verifiable rewards directly from images.
To this end, we revisit self-supervised learning (SSL) in visual domains and reformulate widely-used SSL tasks into a set of verifiable visual puzzles for RL post-training, requiring neither human nor external model supervision.
Training MLLMs on these tasks substantially improves their performance on multimodal understanding and reasoning benchmarks, highlighting the potential of leveraging vision-centric self-supervised tasks for MLLM post-training.
We think this work will provide useful experience in devising effective self-supervised verifiable rewards to enable RL at scale.
Project page: \url{https://github.com/Jiahao000/SSL-R1}.
\end{abstract}


\section{Introduction}
\label{sec:intro}

The success of reinforcement learning (RL) with verifiable rewards (RLVR)~\cite{guo2025deepseek,lambert2024tulu,team2025kimi} in large language models (LLMs)~\cite{brown2020language,openai2023gpt,dubey2024llama3,team2024gemini,meta2025llama,lu2024deepseek,team2025gemma} has motivated researchers to adapt RLVR in the multimodal domain.
Equipped with RLVR, multimodal large language models (MLLMs)~\cite{alayrac2022flamingo,li2022blip,ye2023mplug,liu2023visual,zhu2023minigpt,li2023otter,dai2023instructblip,bai2023qwen,lu2024deepseek,chen2024internvl,bai2025qwen2,zhu2025internvl3} have made significant strides, exhibiting impressive multimodal understanding and reasoning capabilities~\cite{huang2025vision,yu2025perception,meng2025mm,wang2025sota,yuan2025vl}.

Despite their remarkable advancements, as illustrated in Fig.~\ref{fig:teaser}~(a), existing RLVR solutions for post-training MLLMs still face two key challenges:
(\romannumeral1) The rewards are primarily language-centric~\cite{huang2025vision,meng2025mm,wang2025llava,yuan2025vl,wang2025sota}, with dense visual information only being leveraged to extract sparse cues for text-based reasoning.
As a result, they exhibit systematic weaknesses in fine-grained visual understanding.
(\romannumeral2) The training data typically require manual annotations~\cite{liu2025visual,wang2025pixel,wang2025vicrit,liao2025improved}, which is very expensive to obtain and difficult to scale given the increasing complexity and quantity of real-world tasks.
These shortcomings highlight the need for training methods that reinforce vision-centric grounding and reasoning without human supervision.

To this end, we propose SSL-R1, a generic self-supervised post-training framework that derives verifiable rewards directly from input visual signals (see Fig.~\ref{fig:teaser}~(b)). Our method is inspired by commonly used pretext tasks in visual self-supervised learning (SSL), where proxy labels are derived from raw data by corrupting some part of the data. These SSL tasks naturally serve as intrinsic rewards that are suitable for RL-based post-training.
Specifically, our framework defines a suite of verifiable self-supervised visual puzzles based on a single image, including rotation prediction, visual similarity, region inpainting, patch ordering, and geometric correspondence. These visual puzzles require diverse visual understanding and reasoning abilities, with the granularity ranging from the image level to the pixel level.

We evaluate the effectiveness of SSL-R1 on a wide range of multimodal understanding benchmarks (see Fig.~\ref{fig:teaser}~(c)), including fine-grained perception~\cite{tong2024eyes,chen2024we,liu2024mmbench,wang2025divide,wu2024v,zhang2025mme,lai2024lisa,yao2024evaluate}, spatial understanding~\cite{liu2023vsr,jia2025omnispatial,yang2024depth}, and compositional understanding~\cite{thrush2022winoground,dumpala2024sugarcrepe++}. Extensive experimental results demonstrate that our method substantially improves the vision-centric capabilities of MLLMs.

Our main contributions are summarized as follows:

\textbf{1)} We introduce SSL-R1, a generic self-supervised RL post-training framework that derives verifiable rewards intrinsically from images, requiring neither human nor model supervision.

\textbf{2)} We design a diverse set of verifiable self-supervised visual puzzles and contribute a comprehensive empirical study, providing useful insights on what makes for good self-supervised tasks for RL-based post-training.

\textbf{3)} We conduct extensive experiments across multiple evaluation settings. Our SSL-R1 substantially improves the multimodal understanding and reasoning capabilities of MLLMs, demonstrating the potential of leveraging vision-centric self-supervised tasks for MLLM post-training.


\section{Related Work}
\label{sec:work}

\begin{figure*}[t]
    \centering
    \includegraphics[width=\linewidth]{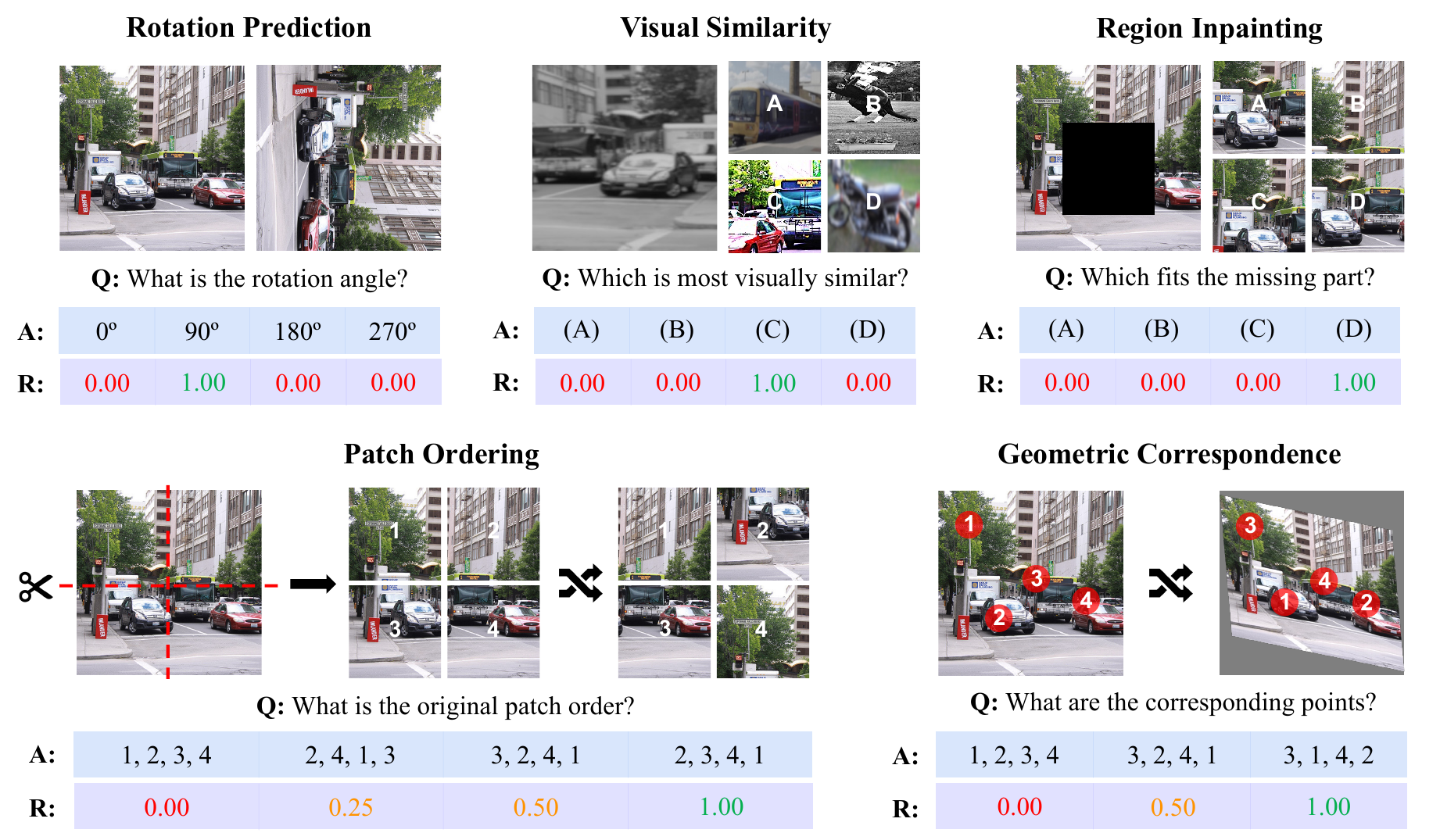}
    \caption{\textbf{Overview of our SSL-R1 tasks.} We design five verifiable self-supervised tasks for RL post-training, ranging from the image level to the pixel level. \emph{Rotation Prediction:} an image is rotated by a certain angle, and the model is tasked with predicting the angle. \emph{Visual Similarity:} two augmented views are cropped from an image, with several additional views from other images, and the task is to select the most visually similar views. \emph{Region Inpainting:} an image is masked by a certain ratio, and the model is required to select the region that best fits the missing part. \emph{Patch Ordering:} an image is partitioned into a grid of patches, shuffled into a sequence, and the model predicts the original patch order. \emph{Geometric Correspondence:} points are sampled from an image, shuffled into a sequence, and the model finds the corresponding point on the transformed image for each point on the original image. Q: Question, A: Answer, R: Reward.}
    \label{fig:pipeline}
\end{figure*}

\begin{table}[t]
\centering
\caption{\textbf{Comparison of key properties with self-supervised RL post-training methods.} Unlike recent or concurrent methods such as Jigsaw-R1~\cite{wang2025jigsaw}, Visual Jigsaw~\cite{wu2025visual}, SSL4RL~\cite{guo2025ssl4rl}, and Spatial-SSRL~\cite{liu2025spatial}, SSL-R1 uniquely satisfies all desired properties: it covers multiple SSL tasks, supports one-time and one-stage training, and is transferable to a broad range of downstream tasks. \textcolor{ForestGreen}{Desired} and \textcolor{magenta}{undesired} properties are shown in \textcolor{ForestGreen}{green} and \textcolor{magenta}{magenta}, respectively.}
\label{tab:related_work}
\resizebox{.48\textwidth}{!}{
\begin{tabular}{lccccc}
\toprule
Desired Properties & \cite{wang2025jigsaw} & \cite{wu2025visual} & \cite{guo2025ssl4rl} & \cite{liu2025spatial} & Ours \\ \midrule
Covers multiple SSL tasks & \textcolor{magenta}{\xmark} & \textcolor{magenta}{\xmark} & \textcolor{ForestGreen}{\ding{51}} & \textcolor{ForestGreen}{\ding{51}} & \textcolor{ForestGreen}{\ding{51}} \\
Supports one-time and one-stage training & \textcolor{ForestGreen}{\ding{51}} & \textcolor{ForestGreen}{\ding{51}} & \textcolor{magenta}{\xmark} & \textcolor{magenta}{\xmark} & \textcolor{ForestGreen}{\ding{51}} \\
Transferable to broad downstream tasks & \textcolor{magenta}{\xmark} & \textcolor{ForestGreen}{\ding{51}} & \textcolor{magenta}{\xmark} & \textcolor{magenta}{\xmark} & \textcolor{ForestGreen}{\ding{51}} \\ \bottomrule
\end{tabular}
}
\end{table}

\noindent\textbf{Self-Supervised Learning.}
Self-supervised learning (SSL) derives supervision from raw data by exploiting their internal priors or structures in the form of a pretext task, which plays a fundamental role in representation learning.
Various pretext tasks have been proposed in the past decade~\cite{doersch2015unsupervised,noroozi2016unsupervised,pathak2016context,gidaris2018unsupervised,caron2018deep,zhan2020online,chen2020simple,he2020momentum,grill2020bootstrap,caron2020unsupervised,chen2020generative,caron2021emerging,xie2021unsupervised,xie2022delving,bao2022beit,he2022masked,zhou2022ibot,xie2023masked,li2023correlational,assran2023self}.
Early examples include context prediction~\cite{doersch2015unsupervised}, jigsaw puzzles~\cite{noroozi2016unsupervised}, inpainting~\cite{pathak2016context}, and rotation~\cite{gidaris2018unsupervised}.
Later approaches based on contrastive learning~\cite{chen2020simple,he2020momentum,caron2021emerging} and masked modeling~\cite{chen2020generative,bao2022beit,he2022masked,zhou2022ibot,assran2023self} have demonstrated impressive scalability, serving as foundations for large-scale pre-training in visual~\cite{fang2023eva,oquab2023dinov2,simeoni2025dinov3}, textual~\cite{devlin2019bert,brown2020language}, and multimodal~\cite{radford2021learning,zhai2023sigmoid,tschannen2025siglip} domains.
However, prior works mostly focus on pre-training, while applying SSL in the post-training stage remains much less explored, especially in the context of MLLMs.

Recently, Jigsaw-R1~\cite{wang2025jigsaw} made the first attempt to use jigsaw puzzles for RL post-training.
Several concurrent works~\cite{wu2025visual,guo2025ssl4rl,liu2025spatial} also attempt to enhance MLLMs via self-supervised post-training.
Specifically, Visual Jigsaw~\cite{wu2025visual} extends jigsaw tasks across three modalities.
SSL4RL~\cite{guo2025ssl4rl} considers multiple SSL tasks while performing training on each downstream task separately.
Spatial-SSRL~\cite{liu2025spatial} mainly targets spatial understanding and requires two-stage training.
In contrast, our SSL-R1 is the \emph{only} method that simultaneously covers multiple SSL tasks, supports one-time and one-stage training, and is transferable to a broad range of downstream tasks.
The comparisons with these works are summarized in Tab.~\ref{tab:related_work}.

\noindent\textbf{Reinforcement Learning with Verifiable Rewards.}
Reinforcement learning (RL) has become a central paradigm to enhance the reasoning capabilities, thanks to the emergence of reasoning models like OpenAI o1~\cite{jaech2024openai}.
The recent success of Deepseek-R1~\cite{guo2025deepseek} has motivated the community to adopt Group Relative Policy Optimization~\cite{shao2024deepseekmath} for rule-based RL, demonstrating the great potential of RL with verifiable rewards (RLVR).
However, most existing RLVR approaches still heavily rely on humans or tool-calling~\cite{huang2025vision,meng2025mm,wang2025llava,yuan2025vl,wang2025sota,liu2025visual,wang2025vicrit,liao2025improved,wen2025reinforcement,wang2025pixel,zheng2025deepeyes} to curate training data for post-training.
Such extrinsic supervision inevitably limits the scalability of RL.
To this end, we introduce a generic self-supervised RL post-training framework that enables intrinsically verifiable supervision at scale, substantially improving the multimodal perception and understanding capabilities of MLLMs.


\section{Methodology}
\label{sec:method}

Our SSL-R1 is a generic self-supervised RL post-training framework. As shown in Fig.~\ref{fig:pipeline}, it includes five self-supervised tasks that target different aspects of visual information, providing comprehensive coverage of vision-centric reasoning capabilities.
In this section, we first introduce related RL preliminaries in Sec.~\ref{subsec:preliminary}. We then detail our SSL-R1 pipeline including task and reward designs in Sec.~\ref{subsec:ssl-r1}.

\subsection{Preliminary}
\label{subsec:preliminary}

\noindent\textbf{Reinforcement Learning with Verifiable Rewards.}
Unlike reinforcement learning from human feedback (RLHF)~\cite{ouyang2022training} that predicts the reward score by a separate reward model trained on preference data, reinforcement learning with verifiable rewards (RLVR)~\cite{guo2025deepseek} determines the reward score directly by pre-defined rules, enhancing models in tasks with objectively verifiable outcomes.
Compared with RLHF, RLVR simplifies the reward mechanism while
maintaining strong alignment with the inherent correctness criteria of the tasks.
Formally, give a reference policy $\pi_0$, RLVR optimizes the policy $\pi_\theta$ with the following objective:
\begin{equation}
\mathop{\text{max}}\limits_{\pi_\theta}\,\mathbb{E}_{y\sim\pi_\theta(x)}[R(x,y)-\beta\,\text{KL}(\pi_\theta(y|x)\,||\,\pi_0(y|x))],
\end{equation}
where $x$ is the input prompt, $y$ is the generated response, $R$ is the verifiable reward, and $\beta$ controls the KL-divergence strength.
Taking the prompt and response pair $(x, y)$ as
inputs, $R$ verifies if $y$ matches the ground-truth answer:
\begin{equation}
R(x,y)=
\begin{cases}
1, \quad \text{if}\;y=\text{ground truth},\\
0, \quad \text{otherwise}.
\end{cases}
\end{equation}

\noindent\textbf{Group Relative Policy Optimization.}
Unlike reinforcement learning algorithms such as PPO~\cite{schulman2017proximal} that require a critic model to evaluate policy performance, Group Relative Policy Optimization (GRPO)~\cite{shao2024deepseekmath} eliminates the need for an additional critic model by comparing groups of candidate responses directly.
Formally, given $L$ sampled responses $\{y_1,y_2,...,y_L\}$ for a prompt $x$ from the current policy, GRPO takes actions based on these responses and obtains rewards $\{r_1,r_2,...,r_L\}$. The advantage of each response is then defined as:
\begin{equation}
A_{i}=\frac{r_i-\text{mean}(\{r_1,r_2,...,r_L\})}{\text{std}(\{r_1,r_2,...,r_L\})},
\end{equation}
where $A_i$ denotes the relative quality of the $i$-th response.
GRPO measures relative performance within the sampled group and encourages the model to favor better answers with a
high reward value.

\subsection{SSL-R1}
\label{subsec:ssl-r1}

\subsubsection{Task Design}

We re-purpose five different self-supervised tasks widely used in the vision literature as examples amenable to being used within an RLVR framework.
We detail each task below.

\noindent\textbf{Rotation Prediction.}
Given an input image $I\in\mathbb{R}^{H\times W\times 3}$, we apply a random rotation clockwise by an angle $y$ to obtain the rotated image $I_\text{rot}$, where $y$ is from a predefined set of angles that are evenly distributed within a 360$^\circ$ range, controlled by a size parameter $S$. Given $I$ and $I_\text{rot}$, the model is prompted to predict the specific rotation angle.
The detailed prompt template for this task is provided in the supplementary material.

\noindent\textbf{Visual Similarity.}
Given an input image $I\in\mathbb{R}^{H\times W\times 3}$, we randomly crop two augmented views from the image, which serve as a reference image $I_\text{ref}$ and a positive image $I_\text{pos}$, respectively. We also crop three additional views from other randomly-selected images, which serve as negative samples $\{I_\text{neg}\}$.
Given the reference image $I_\text{ref}$ and four candidate images $\{I_\text{pos}, \{I_\text{neg}\}\}$ in a multiple-choice format, the model is prompted to select the most visually similar image to the reference image.
The detailed prompt template for this task is provided in the supplementary material.

\noindent\textbf{Region Inpainting.}
Given an input image $I\in\mathbb{R}^{H\times W\times 3}$, we randomly mask a region of the image by a certain masking ratio $\alpha$. The height $h$ and width $w$ of the masked region are uniformly sampled from $\mathcal{U}\left([28, H]\times [28, W]\right)$\footnote{We set the minimal length as 28 to avoid overly small regions.}, whose area should satisfy $h\times w=\alpha\times H\times W$. The top-left corner $(x_0, y_0)$ of the masked region is uniformly sampled from $\mathcal{U}\left([0, W-w]\times [0, H-h]\right)$. We denote the masked input image and the cropped region as $I_\text{mask}$ and $I_\text{crop}$, respectively.
Given $I_\text{mask}$ and four candidate image regions (one ground-truth region plus three random regions) in a multiple-choice format, the model is prompted to select the correct region $I_\text{crop}$ that best fits the missing part of the original image.
The detailed prompt template for this task is provided in the supplementary material.

\noindent\textbf{Patch Ordering.}
Given an input image $I\in\mathbb{R}^{H\times W\times 3}$, we first partition it into $M\times N$ grid of patches, each of size $\frac{H}{M}\times \frac{W}{N}$. We denote the patch sequence as $\mathcal{P}=\left[p_1, p_2, ..., p_{M\times N}\right]$, where $p_i\in\mathbb{R}^{\frac{H}{M}\times \frac{W}{N}\times 3}$.
We then apply a random permutation $\pi:\{1,2,...,M\times N\}\mapsto\{1,2,...,M\times N\}$ to obtain a shuffled patch sequence $\mathcal{P}_{\pi}=[p_{\pi(1)}, p_{\pi(2)}, ..., p_{\pi(M\times N)}]$. Given $\mathcal{P}_{\pi}$, the model is prompted to predict the correct patch ordering $\pi^{-1}=[\pi^{-1}(1), \pi^{-1}(2), ..., \pi^{-1}(M\times N)]$ that recovers the original image.
The detailed prompt template for this task is provided in the supplementary material.

\noindent\textbf{Geometric Correspondence.}
Given an input image $I\in\mathbb{R}^{H\times W\times 3}$, we first randomly select $K$ points\footnote{We ensure each sampled point is separated by at least 40 pixels on the image for diversity.} on the image. We denote the point sequence as $\mathcal{P}=\left[p_1, p_2, ..., p_K\right]$, and each point is annotated with its index in $\mathcal{P}$.
We then apply a random geometric transformation (\eg, affine transformation) to the image and the points to create ground truth for geometric correspondence.
After that, we apply a random permutation $\pi:\{1,2,...,K\}\mapsto\{1,2,...,K\}$ to obtain a shuffled point sequence $\mathcal{P}_{\pi}=[p_{\pi(1)}, p_{\pi(2)}, ..., p_{\pi(K)}]$.
Each point is annotated with its index in $\mathcal{P}_{\pi}$ on the transformed image. For each point on the original image, the model is prompted to find its corresponding point on the transformed image, predicting the correct point ordering $\pi^{-1}=[\pi^{-1}(1), \pi^{-1}(2), ..., \pi^{-1}(K)]$ that recovers $\mathcal{P}$.
The detailed prompt template for this task is provided in the supplementary material.

\begin{table*}[t]
\centering
\caption{\textbf{Ablations on vision-centric multimodal benchmarks.} The baseline model is Qwen2.5-VL-3B. We post-train from Qwen2.5-VL-3B with different SSL tasks, either separately (\emph{Single-Task Post-Training}) or jointly (\emph{Multi-Task Post-Training}). The first five columns denote the SSL tasks in abbreviations, ranked by the average performance (from \emph{left} to \emph{right}): Cor.: \emph{Geometric Correspondence}, Ord.: \emph{Patch Ordering}, Rot.: \emph{Rotation Prediction}, Sim.: \emph{Visual Similarity}, Inp.: \emph{Region Inpainting}. The benchmarks are organized into three categories: \emph{Fine-Grained Perception and Understanding}, \emph{Spatial Understanding}, and \emph{Compositional Understanding}. We report the average accuracy across 13 benchmarks under three categories in the last column. The best results are in \textbf{bold}, and the second-best results are \underline{underlined}.}
\label{tab:ablation}
\resizebox{\textwidth}{!}{
\begin{tabular}{ccccccccccccccccccc}
\toprule
\multicolumn{5}{c}{\multirow{13}{*}{Task}} & \multicolumn{8}{c}{Fine-Grained Perception \& Understanding} & \multicolumn{3}{c}{Spatial Und. (Mono.)} & \multicolumn{2}{c}{Compositional Und.} & \multirow{13}{*}{Avg.} \\ \cmidrule(lr){6-13} \cmidrule(lr){14-16} \cmidrule(lr){17-18}
&  &  &  &  & \rotatebox{90}{MMVP} & \rotatebox{90}{MMStar} & \rotatebox{90}{MMBench} & \rotatebox{90}{HR-Bench-8K} & \rotatebox{90}{V*} & \rotatebox{90}{MME-RealWorld} & \rotatebox{90}{LISA-Grounding} & \rotatebox{90}{OVD-Eval} & \rotatebox{90}{VSR} & \rotatebox{90}{OmniSpatial} & \rotatebox{90}{DA-2K} & \rotatebox{90}{Winoground} & \rotatebox{90}{SugarCrepe++} &  \\ \cmidrule(lr){6-13} \cmidrule(lr){14-16} \cmidrule(lr){17-18}
Cor. & Ord. & Rot. & Sim. & Inp. & test & fine & en\_dev & test & test & lite & test & test & test & test & val & g-acc & test &  \\ \midrule
\multicolumn{19}{l}{\emph{Baseline Model:}} \\
\ding{55} & \ding{55} & \ding{55} & \ding{55} & \ding{55} & 43.33 & 51.49 & 77.15 & 64.12 & 76.44 & 42.68 & 59.59 & 35.21 & 77.31 & 45.27 & 51.89 & 27.75 & 61.05 & 54.87 \\ \midrule
\multicolumn{19}{l}{\emph{Single-Task Post-Training:}} \\
 \ding{51} & \ding{55} & \ding{55} & \ding{55} & \ding{55} & 49.33 & \textbf{56.24} & 77.66 & \textbf{67.63} & 76.96 & 44.09 & 60.13 & 35.35 & 77.81 & \textbf{46.97} & \textbf{58.17} & 29.25 & \textbf{61.43} & \textbf{57.00} \\
 \ding{55} & \ding{51} & \ding{55} & \ding{55} & \ding{55} & \underline{50.00} & 54.00 & 78.26 & 65.00 & \underline{77.49} & 42.63 & \underline{63.03} & \textbf{36.38} & \textbf{78.82} & 46.51 & 54.06 & \underline{30.75} & \underline{61.30} & \underline{56.79} \\
 \ding{55} & \ding{55} & \ding{51} & \ding{55} & \ding{55} & \textbf{52.67} & 53.21 & \underline{78.52} & \underline{65.88} & 76.44 & \underline{44.14} & 60.19 & 35.64 & \underline{77.95} & 46.25 & 53.58 & \textbf{31.25} & 60.94 & 56.67 \\
 \ding{55} & \ding{55} & \ding{55} & \ding{51} & \ding{55} & 47.33 & \underline{54.64} & \textbf{78.69} & 64.88 & 76.96 & \textbf{44.66} & \textbf{63.45} & \underline{36.18} & 77.49 & \underline{46.58} & 53.58 & 29.00 & 60.67 & 56.47 \\
 \ding{55} & \ding{55} & \ding{55} & \ding{55} & \ding{51} & 48.67 & 54.17 & 77.84 & 65.50 & \textbf{81.68} & 42.16 & 59.89 & 36.05 & 76.95 & 45.92 & \underline{54.40} & 27.75 & 61.26 & 56.33 \\
\rowcolor{yellow!10} \multicolumn{5}{c}{$\Delta_{\text{max}}$ (\emph{vs. baseline})} & \textcolor{ForestGreen}{\textbf{+9.34}} & \textcolor{ForestGreen}{\textbf{+4.75}} & \textcolor{ForestGreen}{\textbf{+1.54}} & \textcolor{ForestGreen}{\textbf{+3.51}} & \textcolor{ForestGreen}{\textbf{+5.24}} & \textcolor{ForestGreen}{\textbf{+1.98}} & \textcolor{ForestGreen}{\textbf{+3.86}} & \textcolor{ForestGreen}{\textbf{+1.17}} & \textcolor{ForestGreen}{\textbf{+1.51}} & \textcolor{ForestGreen}{\textbf{+1.70}} & \textcolor{ForestGreen}{\textbf{+6.28}} & \textcolor{ForestGreen}{\textbf{+3.50}} & \textcolor{ForestGreen}{\textbf{+0.38}} & \textcolor{ForestGreen}{\textbf{+2.13}} \\ \midrule
\multicolumn{19}{l}{\emph{Multi-Task Post-Training:}} \\
 \ding{51} & \ding{51} & \ding{55} & \ding{55} & \ding{55} & 51.33 & 54.32 & 78.44 & \underline{68.50} & \underline{79.58} & 43.41 & 61.46 & 36.02 & \textbf{79.13} & \textbf{47.23} & \underline{61.46} & \textbf{30.75} & 61.07 & 57.90 \\
 \ding{51} & \ding{51} & \ding{51} & \ding{55} & \ding{55} & \underline{55.33} & 52.79 & \textbf{78.95} & 67.88 & 79.06 & \textbf{44.66} & \underline{61.58} & \underline{36.12} & 78.63 & \underline{47.03} & 60.30 & 29.25 & \textbf{61.53} & 57.93 \\
 \ding{51} & \ding{51} & \ding{51} & \ding{51} & \ding{55} & \textbf{56.00} & \underline{54.80} & \underline{78.69} & 68.38 & 78.53 & \underline{44.14} & \textbf{62.18} & \textbf{36.46} & \underline{78.95} & 46.90 & 60.20 & \underline{30.25} & \textbf{61.53} & \underline{58.23} \\
 \ding{51} & \ding{51} & \ding{51} & \ding{51} & \ding{51} & 54.67 & \textbf{57.67} & 78.52 & \textbf{68.75} & \textbf{80.63} & 43.67 & \underline{61.58} & 35.60 & 77.68 & 46.84 & \textbf{61.70} & 29.25 & \underline{61.49} & \textbf{58.31} \\
\rowcolor{yellow!10} \multicolumn{5}{c}{$\Delta_{\text{max}}$ (\emph{vs. baseline})} & \textcolor{ForestGreen}{\textbf{+12.67}} & \textcolor{ForestGreen}{\textbf{+6.18}} & \textcolor{ForestGreen}{\textbf{+1.80}} & \textcolor{ForestGreen}{\textbf{+4.63}} & \textcolor{ForestGreen}{\textbf{+4.19}} & \textcolor{ForestGreen}{\textbf{+1.98}} & \textcolor{ForestGreen}{\textbf{+2.59}} & \textcolor{ForestGreen}{\textbf{+1.25}} & \textcolor{ForestGreen}{\textbf{+1.82}} & \textcolor{ForestGreen}{\textbf{+1.96}} & \textcolor{ForestGreen}{\textbf{+9.81}} & \textcolor{ForestGreen}{\textbf{+3.00}} & \textcolor{ForestGreen}{\textbf{+0.48}} & \textcolor{ForestGreen}{\textbf{+3.44}} \\ \bottomrule
\end{tabular}
}
\end{table*}

\subsubsection{Reward Design}

The reward function serves as the primary training signal in RLVR.
Our total reward is the sum of an accuracy reward and a format reward, introduced next.

\noindent\textbf{Accuracy Reward.}
The accuracy reward $R_{\text{acc}}$ assesses the response correctness.
Specifically, for tasks whose answers are a single value (\eg, rotation prediction, visual similarity, and region inpainting), we use a binary reward: $R_{\text{acc}}=1$ if the predicted answer exactly matches the ground truth, and $R_{\text{acc}}=0$ otherwise.
For tasks whose answers are a sequence containing multiple values (\eg, patch ordering and geometric correspondence), we use a graded reward: $R_{\text{acc}}$ is a fractional value between 0 and 1, proportional to the correctly placed indices.

\noindent\textbf{Format Reward.}
The format reward $R_{\text{format}}$ assesses the response format.
Specifically, we require the model to enclose its reasoning process within \texttt{<think> </think>} and the final answer within \texttt{\textbackslash boxed\{\}}. Each tag must appear exactly once and in the correct sequence (the reasoning process before the final answer).
We set $R_{\text{format}}=0.2$ if the output strictly adheres to the prescribed format, and $R_{\text{format}}=0$ otherwise.


\section{Experiments}
\label{sec:exp}

\subsection{Implementation Details}
\label{subsec:details}

\noindent\textbf{SSL Task Setup.}
Unless otherwise specified, we implement the following configurations for each SSL task.
We set $S=4$ for rotation prediction, $\alpha=0.5$ for region inpainting, $M=N=3$ for patch ordering, and $K=6$ for geometric correspondence.
For visual similarity, we apply the standard image augmentations following~\cite{chen2020simple}: a $224\times224$-pixel random resized crop with a random horizontal flip, followed by a random color distortion, random grayscale conversion, and random Gaussian blur.

\noindent\textbf{Training Data.}
We use 118K raw images from COCO~\cite{lin2014microsoft} as our data source to create training data.
Specifically, for each task introduced in Sec.~\ref{subsec:ssl-r1}, we apply the corresponding data augmentation and prompt construction process to generate self-supervised question-answer pairs as training data.
The final dataset consists of 591K question-answer pairs in total, dubbed as SSL-R1-591K, which balances each task in equal proportion and exhibits diverse question formats, including single-value questions, multiple-value questions, and multiple-choice questions.

\noindent\textbf{Baselines.}
We adopt Qwen2.5-VL-3B/7B-Instruct~\cite{bai2025qwen2} as the base MLLMs for all experiments, due to their strong reasoning performances and moderate model sizes.
Apart from the base models, we also compare SSL-R1 with several representative Qwen2.5-VL-based reasoning models that have undergone reasoning-intensive RL post-training in a supervised way, including ThinkLite-VL~\cite{wang2025sota}, VL-Cogito~\cite{yuan2025vl}, LLaVA-Critic-R1~\cite{wang2025llava}.
In addition, to further demonstrate the superiority of our method, we consider a concurrent self-supervised RL post-training method, \eg, Visual Jigsaw~\cite{wu2025visual}, for comparison, with the results reproduced using their official code.

\noindent\textbf{Training Details.}
We optimize the base models using the GRPO algorithm.
During GRPO training, we remove both the KL regularization and the entropy loss following~\cite{wu2025visual}.
For each prompt, we sample 16 responses with a temperature of 1.0.
We train the models with a global batch size of 256 and a learning rate of $1\times10^{-6}$ for either 1000 steps (single-task post-training) or 2000 steps (multi-task post-training).

\begin{figure}[t]
    \centering
    \includegraphics[width=\linewidth]{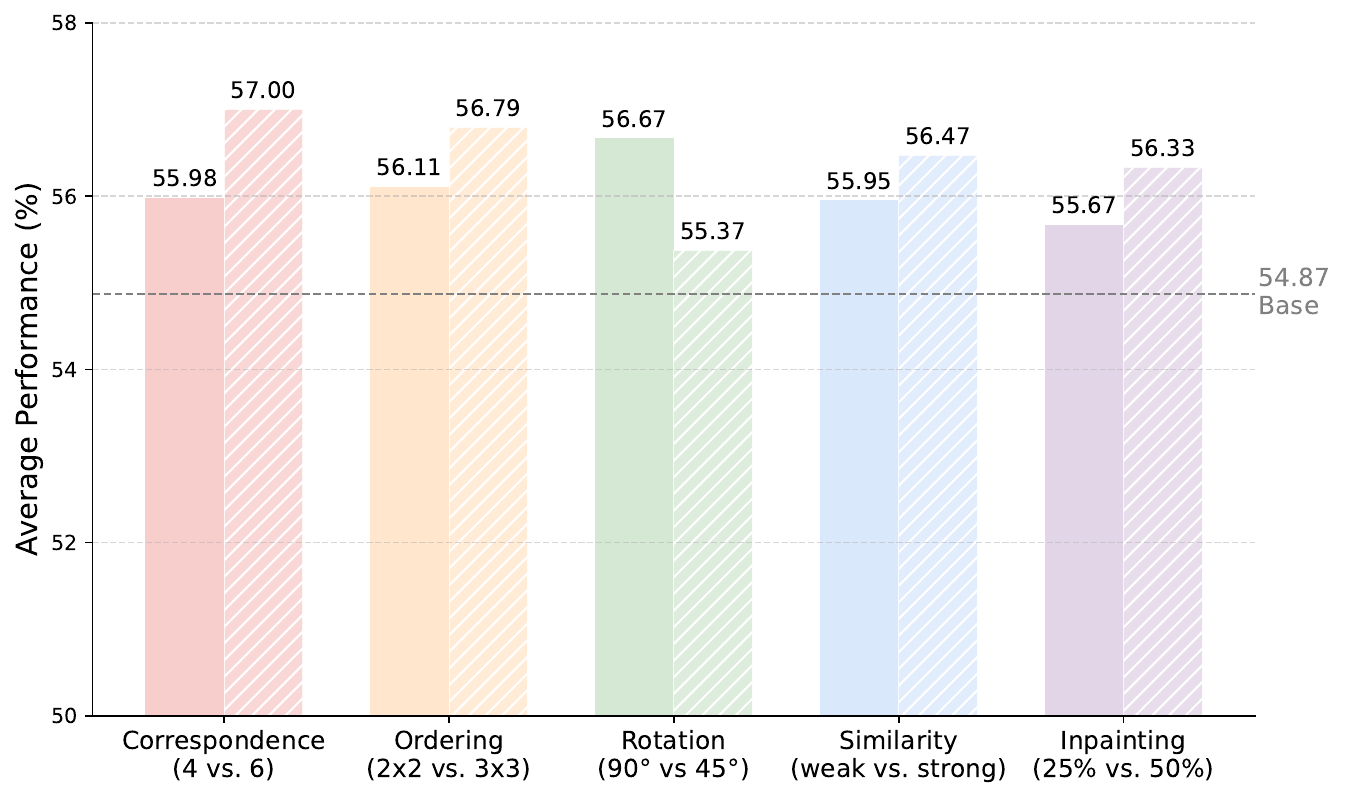}
    \caption{\textbf{Effect of the SSL task difficulty.} We vary the complexity of each task with two difficulty levels, and report the average performance across 13 vision-centric multimodal benchmarks.}
    \label{fig:task_difficulty}
\end{figure}

\noindent\textbf{Benchmarks.}
Following~\cite{wu2025visual}, we evaluate vision-centric capabilities across 13 diverse multimodal benchmarks, covering three main types: (\romannumeral1) Fine-grained perception and understanding: MMVP~\cite{tong2024eyes}, MMStar (fine-grained perception subset)~\cite{chen2024we}, MMBench~\cite{liu2024mmbench}, HR-Bench~\cite{wang2025divide}, V*~\cite{wu2024v}, MME-RealWorld (lite)~\cite{zhang2025mme}, LISA-Grounding~\cite{lai2024lisa}, and OVD-Eval~\cite{yao2024evaluate}; (\romannumeral2) Monocular spatial understanding: VSR~\cite{liu2023vsr}, OminiSpatial~\cite{jia2025omnispatial}, and DA-2K~\cite{yang2024depth}; (\romannumeral3) Compositional visual understanding: Winoground~\cite{thrush2022winoground}, and SugerCrepe++~\cite{dumpala2024sugarcrepe++}.

\subsection{Main Properties}
\label{subsec:properties}

We start by exploring SSL-R1 with each single SSL task, and then combine them for multi-task post-training. We adopt Qwen2.5-VL-3B-Instruct as the base MLLM for all experiments. Several intriguing properties are observed.

\noindent\textbf{Single-Task Post-Training.}
Table~\ref{tab:ablation} (\emph{top section}) studies the effect of individual SSL tasks for RL post-training.
All SSL tasks surpass the baseline on most evaluation benchmarks, demonstrating that each SSL task contributes positively to vision-centric perception and understanding.
Looking at the Avg. column, the greatest gains come from Geometric Correspondence, achieving an average gain of +2.13\% across 13 benchmarks compared to the baseline.
The second best-performing one is Patch Ordering (+1.92\%), with Rotation Prediction coming next (+1.80\%), followed by Visual Similarity (+1.60\%).
Even the least performant Region Inpainting task yields an average gain of +1.46\%.

It is worth mentioning that no single task dominates the others across all benchmarks.
In other words, each task has its own advantages in certain dimensions, and the differences can be very pronounced.
Specifically, Geometric Correspondence excels at fine-grained and high-resolution perception (MMStar, HR-Bench-8K) as well as spatial understanding (DA-2K, OmniSpatial).
In particular, it achieves a significant +6.28\% accuracy improvement for monocular depth estimation on DA-2K.
This is likely due to the fact that finding pixel-level geometric correspondence requires the model to leverage 3D-aware spatial cues for fine-grained and geometric reasoning.
Patch Ordering exhibits balanced performance improvements, ranking within the second place on more than half (7 out of 13) of the benchmarks, suggesting that inferring local patch relations for global spatial layout transfers to downstream tasks broadly.
Rotation Prediction achieves the strongest performance improvement on MMVP (+9.34\%), indicating that identifying image orientations helps understand CLIP-blind visual patterns.
Visual Similarity shines on general visual perception and understanding, achieving the largest number of the best-performing results in the category of fine-grained perception and understanding.
Typically, comparing appearance similarity among views requires strong (multi-object) semantic reasoning, which is beneficial for general VQA tasks.
Region Inpainting stands out on fine-grained visual search (V*), yielding a promising +5.24\% accuracy gain, demonstrating that inpainting local regions attends to pixel-level details and benefits contextual understanding and reasoning.

\begin{figure}[t]
    \centering
    \includegraphics[width=\linewidth]{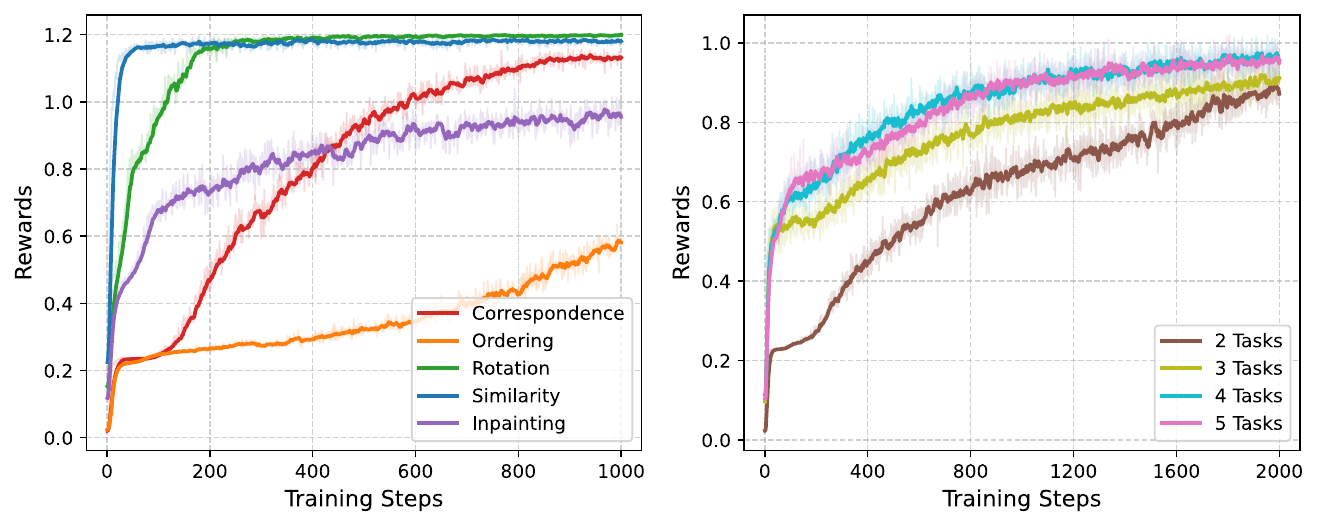}
    \caption{\textbf{The training dynamics of SSL-R1.} \emph{Left:} Single-task rewards. \emph{Right:} Multi-task rewards. All curves are exponentially smoothed for visualization.}
    \label{fig:training_dynamics}
\end{figure}

\begin{table*}[t]
\centering
\caption{\textbf{Performance of Qwen2.5-VL-based models on vision-centric multimodal benchmarks.} All models are evaluated on 13 benchmarks, and the average results across them are provided in the last column.}
\label{tab:final}
\resizebox{\textwidth}{!}{
\begin{tabular}{lcccccccccccccc}
\toprule
 & \multicolumn{8}{c}{Fine-Grained Perception \& Understanding} & \multicolumn{3}{c}{Spatial Und. (Mono.)} & \multicolumn{2}{c}{Compositional Und.} &  \\ \cmidrule(lr){2-9} \cmidrule(lr){10-12} \cmidrule(lr){13-14}
 & \rotatebox{90}{MMVP} & \rotatebox{90}{MMStar} & \rotatebox{90}{MMBench} & \rotatebox{90}{HR-Bench-8K} & \rotatebox{90}{V*} & \rotatebox{90}{MME-RealWorld} & \rotatebox{90}{LISA-Grounding} & \rotatebox{90}{OVD-Eval} & \rotatebox{90}{VSR} & \rotatebox{90}{OmniSpatial} & \rotatebox{90}{DA-2K} & \rotatebox{90}{Winoground} & \rotatebox{90}{SugarCrepe++} &  \\ \cmidrule(lr){2-9} \cmidrule(lr){10-12} \cmidrule(lr){13-14}
\multirow{-3}{*}{Model} & test & fine & en\_dev & test & test & lite & test & test & test & test & val & g-acc & test & \multirow{-3}{*}{Avg.} \\ \midrule
\multicolumn{15}{l}{\emph{Representative Reasoning Models:}} \\
ThinkLite-VL-7B~\cite{wang2025sota} & 55.33 & 59.95 & 84.19 & 68.12 & 76.96 & 46.17 & 73.70 & 35.78 & 78.09 & 42.60 & 58.46 & 35.25 & 61.49 & 59.70 \\
VL-Cogito-7B~\cite{yuan2025vl} & 55.33 & 56.64 & 82.98 & 69.62 & 79.58 & 47.63 & 72.26 & 35.78 & 79.82 & 44.29 & 56.43 & 38.25 & 63.59 & 60.17 \\
LLaVA-Critic-R1-7B~\cite{wang2025llava} & 53.33 & 57.80 & 83.16 & 67.50 & 78.01 & 45.18 & 68.52 & 35.28 & 78.50 & 42.73 & 53.82 & 34.75 & 61.93 & 58.50 \\ \midrule\midrule
\multicolumn{15}{l}{\emph{Baseline vs. Ours (3B):}} \\
Qwen2.5-VL-3B~\cite{bai2025qwen2} & 43.33 & 51.49 & 77.15 & 64.12 & 76.44 & 42.68 & 59.59 & 35.21 & 77.31 & 45.27 & 51.89 & 27.75 & 61.05 & 54.87 \\
+ Visual Jigsaw~\cite{wu2025visual} & 50.67 & 53.88 & 78.44 & 66.50 & 78.53 & 42.26 & 63.03 & 36.43 & 78.63 & 46.12 & 55.42 & 27.25 & 60.21 & 56.72 \\
\rowcolor{yellow!10} + SSL-R1 & 54.67 & 57.67 & 78.52 & 68.75 & 80.63 & 43.67 & 61.58 & 35.60 & 77.68 & 46.84 & 61.70 & 29.25 & 61.49 & 58.31 \\
\rowcolor{yellow!10} $\Delta$ (\emph{vs. baseline}) & \textcolor{ForestGreen}{\textbf{+11.34}} & \textcolor{ForestGreen}{\textbf{+6.18}} & \textcolor{ForestGreen}{\textbf{+1.37}} & \textcolor{ForestGreen}{\textbf{+4.63}} & \textcolor{ForestGreen}{\textbf{+4.19}} & \textcolor{ForestGreen}{\textbf{+0.99}} & \textcolor{ForestGreen}{\textbf{+1.99}} & \textcolor{ForestGreen}{\textbf{+0.39}} & \textcolor{ForestGreen}{\textbf{+0.37}} & \textcolor{ForestGreen}{\textbf{+1.57}} & \textcolor{ForestGreen}{\textbf{+9.81}} & \textcolor{ForestGreen}{\textbf{+1.50}} & \textcolor{ForestGreen}{\textbf{+0.44}} & \textcolor{ForestGreen}{\textbf{+3.44}} \\ \midrule
\multicolumn{15}{l}{\emph{Baseline vs. Ours (7B):}} \\
Qwen2.5-VL-7B~\cite{bai2025qwen2} & 54.66 & 59.75 & 83.33 & 67.38 & 76.96 & 43.41 & 71.89 & 35.07 & 77.68 & 42.66 & 54.45 & 37.00 & 61.59 & 58.91 \\
+ Visual Jigsaw~\cite{wu2025visual} & 59.33 & 60.86 & 83.68 & 70.88 & 82.20 & 47.58 & 74.79 & 36.41 & 80.27 & 45.21 & 59.82 & 39.25 & 63.04 & 61.79 \\
\rowcolor{yellow!10} + SSL-R1 & 60.00 & 62.81 & 84.11 & 71.75 & 81.68 & 49.87 & 74.07 & 36.50 & 81.05 & 45.40 & 63.54 & 41.25 & 63.67 & 62.75 \\
\rowcolor{yellow!10} $\Delta$ (\emph{vs. baseline}) & \textcolor{ForestGreen}{\textbf{+5.34}} & \textcolor{ForestGreen}{\textbf{+3.06}} & \textcolor{ForestGreen}{\textbf{+0.78}} & \textcolor{ForestGreen}{\textbf{+4.37}} & \textcolor{ForestGreen}{\textbf{+4.72}} & \textcolor{ForestGreen}{\textbf{+6.46}} & \textcolor{ForestGreen}{\textbf{+2.18}} & \textcolor{ForestGreen}{\textbf{+1.43}} & \textcolor{ForestGreen}{\textbf{+3.37}} & \textcolor{ForestGreen}{\textbf{+2.74}} & \textcolor{ForestGreen}{\textbf{+9.09}} & \textcolor{ForestGreen}{\textbf{+4.25}} & \textcolor{ForestGreen}{\textbf{+2.08}} & \textcolor{ForestGreen}{\textbf{+3.84}} \\ \bottomrule
\end{tabular}
}
\end{table*}

\noindent\textbf{Multi-Task Post-Training.}
Given that each SSL task improves vision-centric perception and understanding in different ways, it is interesting to ask whether combining multiple SSL tasks further benefits.
We study this by performing multi-task post-training.
We start from the single task formulation and gradually add each new task based on the average performance ranking in the single-task post-training.
As shown in Tab.~\ref{tab:ablation} (\emph{bottom section}), combining two best-performing tasks already improves individual tasks by clear margins (+0.90\% vs. Geometric Correspondence, +1.11\% vs. Patch Ordering).
Adding more tasks leads to further gains, with the best performance achieved by combining \emph{all five tasks}, which outperforms the best-performing single task by 1.31\% on average.
Since each task encourages the model to acquire different visual reasoning abilities, they could be complementary to each other.
This task synergy highlights the importance of devising diverse SSL objectives for multimodal perception and understanding.

On individual benchmarks, we have two observations.
First, combining multiple SSL tasks does not always benefit.
For example, on LISA-Grounding, the best performance in the multi-task setting is lower than that in the single-task setting (62.18\% vs. 63.45\%).
Second, the best performance is not always achieved by combining all SSL tasks.
For instance, on OmniSpatial, we obtain the best performance by combining two instead of five tasks (47.23\% vs. 46.84\%).
Adding more tasks beyond two continuously degrades the performance.
Both phenomena indicate that different tasks may conflict in certain scenarios, causing a negative transfer.
In Sec.~\ref{subsec:analysis}, we further analyze the transfer effect among different SSL tasks, revealing that the task synergy and conflict indeed exist.

Overall, in practice, we recommend using diverse SSL task combinations for holistic multimodal perception and understanding.
Nevertheless, if one prefers optimizing certain downstream capabilities (\eg, visual search), prioritizing corresponding SSL tasks (\eg, Region Inpainting) for post-training would be a better choice.

\noindent\textbf{Task Difficulty.}
We study how the difficulty of SSL tasks affects model performance by varying the complexity of each task with two difficulty levels.
Specifically, for Geometric Correspondence, we reduce the number of sampled points from 6 to 4; for Patch Ordering, we reduce the grid size from $3\times 3$ to $2\times 2$; for Rotation Prediction, we reduce the angle interval from 90$^\circ$ to 45$^\circ$; for Visual Similarity, we reduce the augmentation strength from strong to weak; for Region Inpainting, we reduce the masking ratio from 0.5 to 0.25.
Figure~\ref{fig:task_difficulty} reports the average performance across all benchmarks.
All configurations yield performance improvements over the base model, while a harder task generally leads to more gains, indicating that a higher degree of difficulty provides a stronger supervision signal for downstream reasoning.
We also notice that increasing the task difficulty of Rotation Prediction degrades the performance.
A possible explanation is that using an overly fine-grained rotation operation might induce overfitting to the SSL objective, leading to a negative transfer to downstream tasks.

\noindent\textbf{Training Dynamics.}
Figure~\ref{fig:training_dynamics} shows how our SSL-R1 evolves during training.
Generally, the rewards gradually increase with the training steps, while different tasks exhibit different trends.
Specifically, Visual Similarity and Rotation Prediction rapidly increase in early training stages and finally converge to near-perfect accuracies.
Region Inpainting has a lower reward initially and increases gradually at a steady speed.
Geometric Correspondence and Patch Ordering have the lowest initial rewards and evolve slowly at the beginning, while the growth rates get faster in later stages.
For multi-task post-training, adding more tasks generally leads to increased rewards, likely because multiple tasks synergize and make the overall learning process easier.

\begin{table*}[t]
\centering
\caption{\textbf{Evaluation on self-supervised tasks.} Each task has two difficulty levels. We also report the average accuracy across all tasks in the last column. The best results are in \textbf{bold}, and the second-best results are \underline{underlined}. $\Delta$ denotes the gains relative to Qwen2.5-VL-3B.}
\label{tab:ssl}
\resizebox{\textwidth}{!}{
\begin{tabular}{llccccccccccc}
\toprule
\multirow{2}{*}{Type} & \multirow{2}{*}{Model} & \multicolumn{2}{c}{Correspondence} & \multicolumn{2}{c}{Ordering} & \multicolumn{2}{c}{Rotation} & \multicolumn{2}{c}{Similarity} & \multicolumn{2}{c}{Inpainting} & \multirow{2}{*}{Avg.} \\ \cmidrule(lr){3-4}\cmidrule(lr){5-6}\cmidrule(lr){7-8}\cmidrule(lr){9-10}\cmidrule(lr){11-12}
 &  & 4 points & 6 points & 2$\times$2 & 3$\times$3 & 4$\times$90$^\circ$ & 8$\times$45$^\circ$ & weak aug. & strong aug. & 25\% & 50\% &  \\ \midrule
\multirow{2}{*}{Baseline} & Random & 4.17 & 0.14 & 4.17 & 0.00 & 25.00 & 12.50 & 25.00 & 25.00 & 25.00 & 25.00 & 14.60 \\
 & Qwen2.5-VL-3B~\cite{bai2025qwen2} & 0.28 & 0.02 & 3.16 & 0.00 & 18.64 & 7.64 & 50.18 & 47.32 & 20.96 & 17.66 & 16.59 \\ \midrule
 & Correspondence & \textbf{93.50} & \textbf{92.18} & 6.66 & 0.00 & 26.06 & 12.30 & 96.38 & 93.38 & 25.08 & 23.12 & \underline{46.87} \\
\rowcolor{yellow!10} \cellcolor{white} & $\Delta$ & \textcolor{ForestGreen}{\textbf{+93.22}} & \textcolor{ForestGreen}{\textbf{+92.16}} & \textcolor{ForestGreen}{\textbf{+3.50}} & \eqlval{\textbf{+0.00}} & \textcolor{ForestGreen}{\textbf{+7.42}} & \textcolor{ForestGreen}{\textbf{+4.66}} & \textcolor{ForestGreen}{\textbf{+46.20}} & \textcolor{ForestGreen}{\textbf{+46.06}} & \textcolor{ForestGreen}{\textbf{+4.12}} & \textcolor{ForestGreen}{\textbf{+5.46}} & \textcolor{ForestGreen}{\textbf{+30.28}} \\ \cmidrule(l){2-13}
 & Ordering & 4.92 & 0.14 & \textbf{56.72} & \textbf{31.16} & 27.74 & 14.82 & 96.04 & 92.56 & 30.28 & 25.84 & 38.02 \\
\rowcolor{yellow!10} \cellcolor{white} & $\Delta$ & \textcolor{ForestGreen}{\textbf{+4.64}} & \textcolor{ForestGreen}{\textbf{+0.12}} & \textcolor{ForestGreen}{\textbf{+53.56}} & \textcolor{ForestGreen}{\textbf{+31.16}} & \textcolor{ForestGreen}{\textbf{+9.10}} & \textcolor{ForestGreen}{\textbf{+7.18}} & \textcolor{ForestGreen}{\textbf{+45.86}} & \textcolor{ForestGreen}{\textbf{+45.24}} & \textcolor{ForestGreen}{\textbf{+9.32}} & \textcolor{ForestGreen}{\textbf{+8.18}} & \textcolor{ForestGreen}{\textbf{+21.43}} \\ \cmidrule(l){2-13}
 & Rotation & 2.56 & 0.12 & 6.72 & 0.00 & \underline{99.70} & \underline{49.82} & 93.18 & 88.72 & 32.56 & 25.22 & 39.86 \\
\rowcolor{yellow!10} \cellcolor{white} & $\Delta$ & \textcolor{ForestGreen}{\textbf{+2.28}} & \textcolor{ForestGreen}{\textbf{+0.10}} & \textcolor{ForestGreen}{\textbf{+3.56}} & \eqlval{\textbf{+0.00}} & \textcolor{ForestGreen}{\textbf{+81.06}} & \textcolor{ForestGreen}{\textbf{+42.18}} & \textcolor{ForestGreen}{\textbf{+43.00}} & \textcolor{ForestGreen}{\textbf{+41.40}} & \textcolor{ForestGreen}{\textbf{+11.60}} & \textcolor{ForestGreen}{\textbf{+7.56}} & \textcolor{ForestGreen}{\textbf{+23.27}} \\ \cmidrule(l){2-13}
 & Similarity & 5.00 & 0.30 & 6.54 & 0.00 & 29.36 & 14.92 & \underline{98.82} & \textbf{98.06} & 10.82 & 13.60 & 27.74 \\
\rowcolor{yellow!10} \cellcolor{white} & $\Delta$ & \textcolor{ForestGreen}{\textbf{+4.72}} & \textcolor{ForestGreen}{\textbf{+0.28}} & \textcolor{ForestGreen}{\textbf{+3.38}} & \eqlval{\textbf{+0.00}} & \textcolor{ForestGreen}{\textbf{+10.72}} & \textcolor{ForestGreen}{\textbf{+7.28}} & \textcolor{ForestGreen}{\textbf{+48.64}} & \textcolor{ForestGreen}{\textbf{+50.74}} & \negval{\textbf{-10.14}} & \negval{\textbf{-4.06}} & \textcolor{ForestGreen}{\textbf{+11.15}} \\ \cmidrule(l){2-13}
 & Inpainting & 4.80 & 0.14 & 7.82 & 0.00 & 32.18 & 12.24 & 88.28 & 84.98 & \underline{79.72} & \textbf{74.08} & 38.42 \\
\rowcolor{yellow!10} \cellcolor{white} & $\Delta$ & \textcolor{ForestGreen}{\textbf{+4.52}} & \textcolor{ForestGreen}{\textbf{+0.12}} & \textcolor{ForestGreen}{\textbf{+4.66}} & \eqlval{\textbf{+0.00}} & \textcolor{ForestGreen}{\textbf{+13.54}} & \textcolor{ForestGreen}{\textbf{+4.60}} & \textcolor{ForestGreen}{\textbf{+38.10}} & \textcolor{ForestGreen}{\textbf{+37.66}} & \textcolor{ForestGreen}{\textbf{+58.76}} & \textcolor{ForestGreen}{\textbf{+56.42}} & \textcolor{ForestGreen}{\textbf{+21.83}} \\ \cmidrule(l){2-13}
\multirow{-12}{*}{SSL-R1} & Combined & \underline{92.32} & \underline{89.94} & \underline{54.00} & \underline{10.14} & \textbf{99.74} & \textbf{50.64} & \textbf{99.10} & \underline{97.72} & \textbf{80.00} & \underline{68.22} & \textbf{74.18} \\
\rowcolor{yellow!10} \cellcolor{white} & $\Delta$ & \textcolor{ForestGreen}{\textbf{+92.04}} & \textcolor{ForestGreen}{\textbf{+89.92}} & \textcolor{ForestGreen}{\textbf{+50.84}} & \textcolor{ForestGreen}{\textbf{+10.14}} & \textcolor{ForestGreen}{\textbf{+81.10}} & \textcolor{ForestGreen}{\textbf{+43.00}} & \textcolor{ForestGreen}{\textbf{+48.92}} & \textcolor{ForestGreen}{\textbf{+50.40}} & \textcolor{ForestGreen}{\textbf{+59.04}} & \textcolor{ForestGreen}{\textbf{+50.56}} & \textcolor{ForestGreen}{\textbf{+57.59}} \\ \bottomrule
\end{tabular}
}
\end{table*}

\begin{table*}[ht]
\centering
\caption{\textbf{Performance of SSL-R1 on reasoning models.} The baseline model is ThinkLite-VL-7B. We evaluate on both vision-centric and reasoning-centric multimodal benchmarks. The average results are provided in the last column.}
\label{tab:reasoning}
\resizebox{\textwidth}{!}{
\begin{tabular}{llllllll}
\toprule
\multirow{2}{*}{Model} & \multicolumn{1}{c}{Vision-Centric} & \multicolumn{1}{c}{MathVista} & \multicolumn{1}{c}{MathVision} & \multicolumn{1}{c}{MathVerse} & \multicolumn{1}{c}{MMMU} & \multicolumn{1}{c}{EMMA} & \multicolumn{1}{c}{\multirow{2}{*}{Avg.}} \\ \cmidrule(lr){2-7}
 & \multicolumn{1}{c}{avg} & \multicolumn{1}{c}{testmini} & \multicolumn{1}{c}{testmini} & \multicolumn{1}{c}{testmini} & \multicolumn{1}{c}{val} & \multicolumn{1}{c}{mini} & \multicolumn{1}{c}{} \\ \midrule
ThinkLite-VL-7B~\cite{wang2025sota} & 59.70 & 75.20 & 30.92 & 50.76 & 55.11 & 26.75 & 49.74 \\
+ Visual Jigsaw~\cite{wu2025visual} & 61.60 \textcolor{ForestGreen}{(\textbf{+1.90})} & 75.10 \negval{(\textbf{-0.10})} & 35.20 \textcolor{ForestGreen}{(\textbf{+4.28})} & 50.50 \negval{(\textbf{-0.26})} & 54.22 \negval{(\textbf{-0.89})} & 29.00 \textcolor{ForestGreen}{(\textbf{+2.25})} & 50.94 \textcolor{ForestGreen}{(\textbf{+1.20})} \\
\rowcolor{yellow!10} + SSL-R1 & 62.35 \textcolor{ForestGreen}{(\textbf{+2.65})} & 76.10 \textcolor{ForestGreen}{(\textbf{+0.90})} & 36.25 \textcolor{ForestGreen}{(\textbf{+5.33})} & 51.00 \textcolor{ForestGreen}{(\textbf{+0.24})} & 55.78 \textcolor{ForestGreen}{(\textbf{+0.67})} & 29.34 \textcolor{ForestGreen}{(\textbf{+2.59})} & 51.80 \textcolor{ForestGreen}{(\textbf{+2.06})} \\ \bottomrule
\end{tabular}
}
\end{table*}

\subsection{Comparison with Previous Methods}
\label{subsec:comparison}

In Tab.~\ref{tab:final}, we compare Qwen2.5-VL-based models on 13 diverse vision-centric multimodal benchmarks, spanning fine-grained perception and understanding, spatial understanding, and compositional understanding.
Our final model, SSL-R1, achieves consistent improvements on Qwen2.5-VL across all benchmarks for both 3B and 7B model scales.
Specifically, SSL-R1-3B achieves an average gain of +3.44\%, with particularly noteworthy improvements on MMVP (+11.34\%), DA-2K (+9.81\%), and MMStar (+6.18\%).
Scaling up to SSL-R1-7B further yields an average gain of +3.84\%.

Note that our self-supervised 7B model significantly outperforms prior reasoning models (\eg, ThinkLite-VL, VL-Cogito, LLaVA-Critic-R1) that undergo reasoning-intensive RL post-training in a supervised way.
This demonstrates the great potential of self-supervised verifiable rewards for RL post-training.
In addition, our SSL-R1 also outperforms concurrent self-supervised RL post-training works like Visual Jigsaw by a clear margin, further proving the superiority of our method.

\subsection{Further Analysis}
\label{subsec:analysis}

\noindent\textbf{How Do MLLMs Perform on SSL Tasks?}
In Sec.~\ref{subsec:properties}, we have studied the effect of transferring SSL tasks to various downstream tasks.
Since the designed SSL tasks are 100\% verifiable, we are further interested in investigating how well MLLMs perform on these SSL tasks and the transfer effect among them.
As shown in Tab.~\ref{tab:ssl}, existing MLLMs like Qwen2.5-VL achieve near-random average accuracy on these SSL benchmarks before post-training. In most cases, it is even worse than the random baseline on individual benchmarks.
After post-training with every single SSL task, we observe a significant accuracy improvement on the corresponding task type.
Interestingly, post-training on one task can potentially benefit the others.
For example, the accuracy on Visual Similarity gets significantly improved when the model is trained with any other tasks, suggesting that all the other tasks have a positive transfer effect for Visual Similarity, likely because all of them require the model to reason visual semantics that are beneficial for appearance similarity.
However, we find some hard-transfer tasks (\eg, Geometric Correspondence and Patch Ordering), where training with other tasks leads to marginal or no performance improvements.
Since Geometric Correspondence and Patch Ordering are the two hardest tasks, where Qwen2.5-VL achieves near-zero accuracies, post-training with other easier SSL tasks cannot substantially benefit.
Besides, we also notice a negative transfer case: training with Visual Similarity degrades the accuracy on Region Inpainting, likely because the intensive augmentation in Visual Similarity could be detrimental for recognizing pixel-level details in Region Inpainting.
Nevertheless, combining multiple tasks for post-training consistently improves the accuracy on all tasks, sometimes even better than those trained with the corresponding tasks, \eg, Rotation Prediction (4$\times$90$^\circ$, 8$\times$45$^\circ$), Visual Similarity (weak aug.), and Region Inpainting (25\%).
This further demonstrates the synergy of multiple tasks during post-training.

\begin{table}[t]
\centering
\caption{\textbf{Performance of SSL-R1 on other base models.} We take MiMo-VL as an example, and report the average accuracy across 13 vision-centric multimodal benchmarks.}
\label{tab:mimo}
\begin{tabular}{ll}
\toprule
Model & Vision-Centric Avg. \\ \midrule
MiMo-VL-7B-SFT~\cite{xiaomi2506mimo} & 63.77 \\
+ Visual Jigsaw~\cite{wu2025visual} & 65.14 \textcolor{ForestGreen}{(\textbf{+1.37})} \\
\rowcolor{yellow!10} + SSL-R1 & 66.65 \textcolor{ForestGreen}{(\textbf{+2.88})} \\ \bottomrule
\end{tabular}
\end{table}

\noindent\textbf{How Does SSL-R1 Perform on Reasoning Models?}
We further examine whether SSL-R1 remains effective on reasoning models that have undergone intensive RL post-training.
Table~\ref{tab:reasoning} presents the results of applying SSL-R1 to a strong multimodal reasoning model, \ie, ThinkLite-VL.
Apart from previous vision-centric multimodal benchmarks, we also evaluate the models on reasoning-centric multimodal benchmarks, including MathVista~\cite{lu2024mathvista}, MathVision~\cite{wang2024measuring}, MathVerse~\cite{zhang2024mathverse}, MMMU~\cite{yue2024mmmu}, and EMMA~\cite{hao2025can}.
Despite the reasoning-intensive post-training ThinkLite-VL has undergone, SSL-R1 still yields an average gain of 2.06\%, further boosting its visual perception and understanding while preserving strong reasoning capabilities.
It should be well noted that Visual Jigsaw degrades the performance on several reasoning-heavy benchmarks (\eg, MathVista, MathVerse, MMMU), whereas our SSL-R1 consistently outperforms the baseline on all benchmarks.

\noindent\textbf{How Does SSL-R1 Perform on Other Base Models?}
Our results thus far are based on Qwen2.5-VL architecture and its post-trained models. Actually, our designed SSL tasks are agnostic to architectures. To examine the general effectiveness of our method on other base models, we further apply SSL-R1 on a stronger base model, \eg, MiMo-VL-7B-SFT-2508~\cite{xiaomi2506mimo}. As shown in Tab.~\ref{tab:mimo}, SSL-R1 again significantly outperforms the baseline and surpasses Visual Jigsaw by a clear margin.


\section{Conclusion}
\label{sec:conclusion}

We have introduced SSL-R1, a generic self-supervised RL post-training framework that derives intrinsically verifiable rewards from input images.
SSL-R1 is vision-centric, cost-effective, and scalable, requiring neither human nor external model supervision.
Our extensive experiments demonstrate that SSL-R1 achieves substantial and consistent improvements on a wide range of multimodal understanding and reasoning benchmarks.
We hope our explorations can pave the way for enabling RLVR at scale.

{
    \small
    \bibliographystyle{ieeenat_fullname}
    \bibliography{main}
}


\appendix
\section*{Appendix}

In this supplementary material, we provide the detailed prompt templates (Sec.~\ref{sec:templates}), self-supervised task examples (Sec.~\ref{sec:task_examples}), and qualitative benchmark results (Sec.~\ref{sec:qualitative_results}) to complement the main paper.

\section{Prompt Templates}
\label{sec:templates}

\begin{figure*}[ht]
    \centering
    \includegraphics[width=\linewidth]{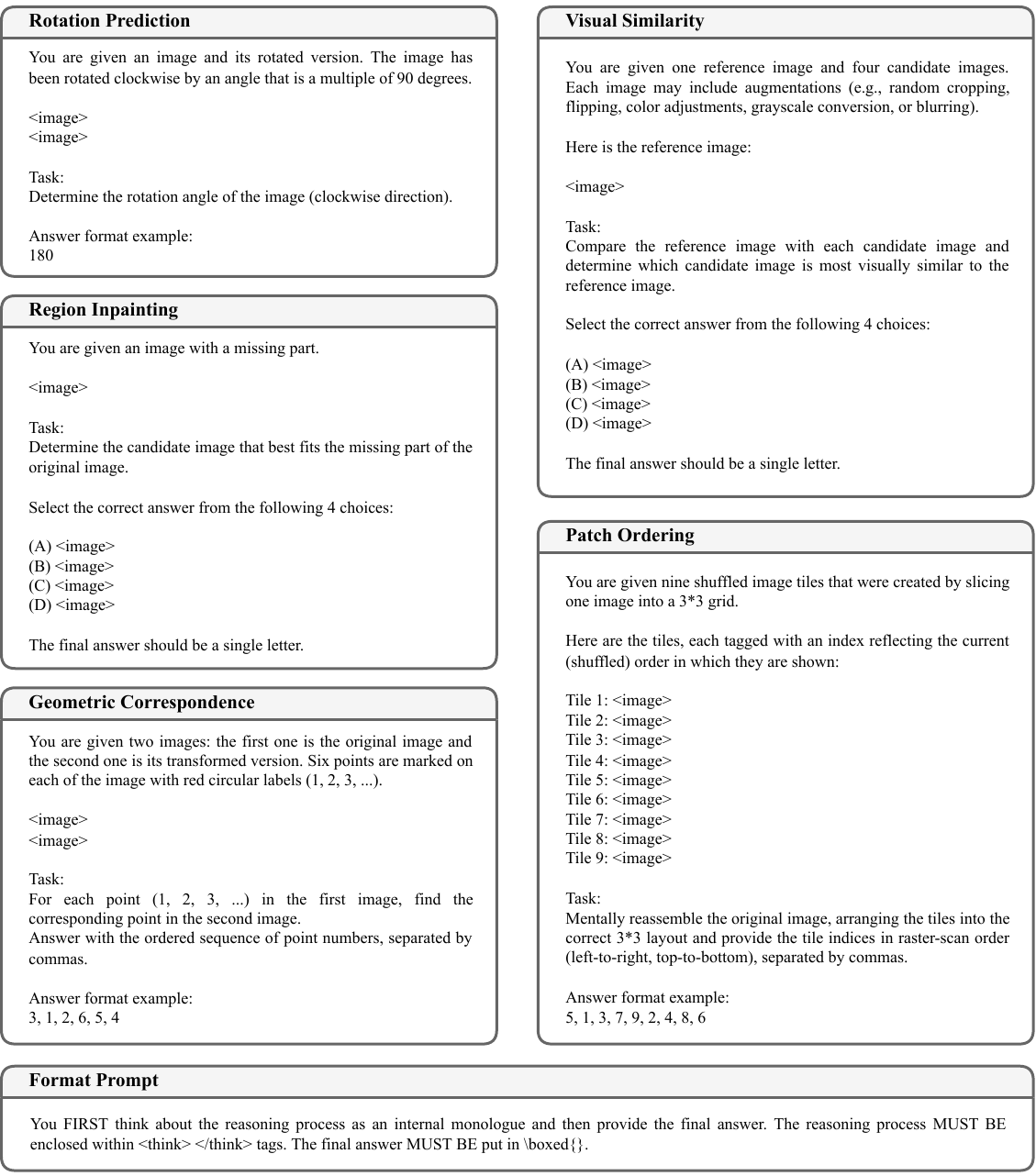}
    \caption{\textbf{Prompt templates for five self-supervised tasks.} A format prompt (\emph{bottom}) is appended at the end of each task prompt to enable the reasoning process.}
    \label{fig:prompt_template}
\end{figure*}

We provide prompt templates for the five self-supervised tasks in Fig.~\ref{fig:prompt_template}.
Note that we also append a format prompt at the end of each task prompt to enable the model to generate the reasoning content.

\section{Task Examples}
\label{sec:task_examples}

\begin{figure*}[ht]
    \centering
    \includegraphics[width=\linewidth]{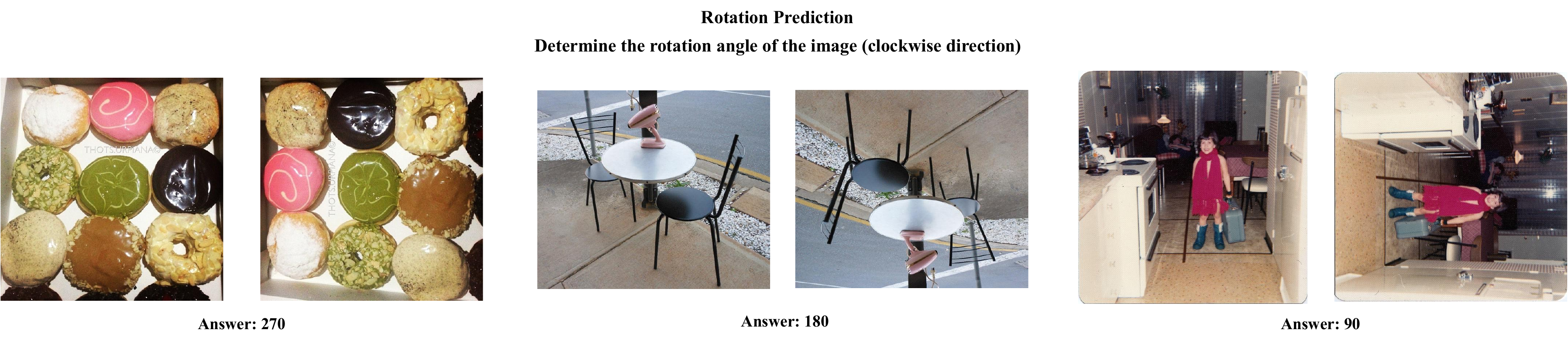}
    \caption{\textbf{Examples of the Rotation Prediction task.} The ground-truth answer for each example is provided at the bottom.}
    \label{fig:task_rotation}
\end{figure*}

\begin{figure*}[ht]
    \centering
    \includegraphics[width=\linewidth]{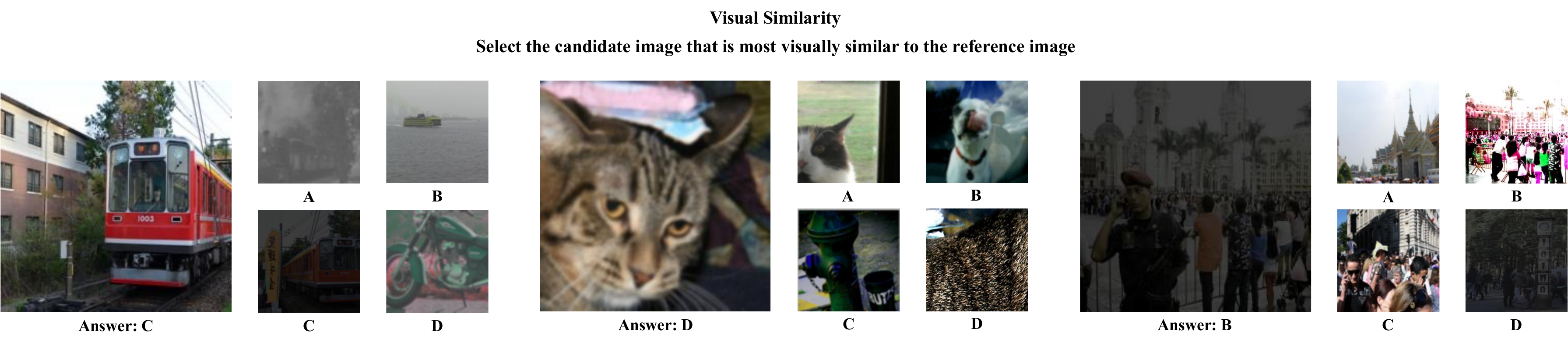}
    \caption{\textbf{Examples of the Visual Similarity task.} The ground-truth answer for each example is provided at the bottom.}
    \label{fig:task_similarity}
\end{figure*}

\begin{figure*}[ht]
    \centering
    \includegraphics[width=\linewidth]{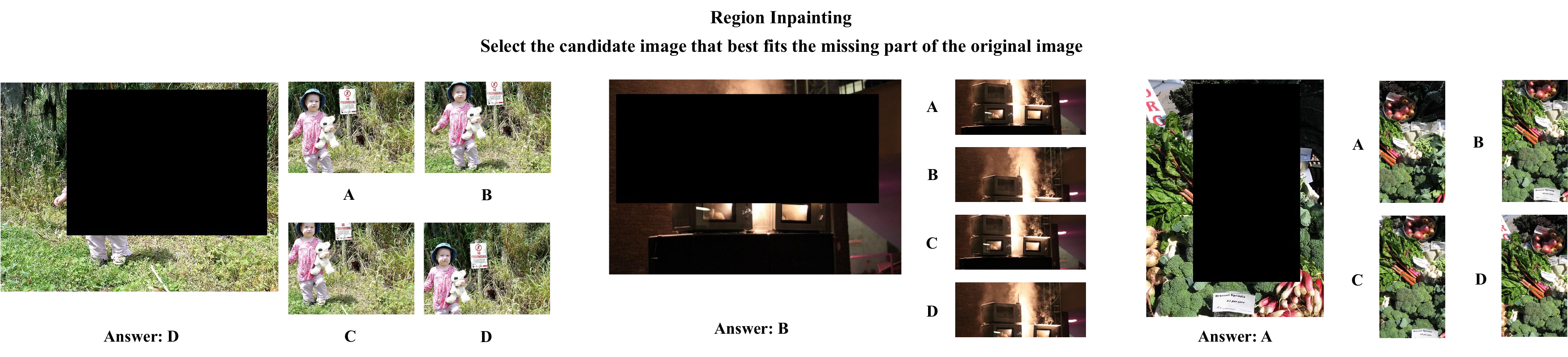}
    \caption{\textbf{Examples of the Region Inpainting task.} The ground-truth answer for each example is provided at the bottom.}
    \label{fig:task_inpainting}
\end{figure*}

\begin{figure*}[ht]
    \centering
    \includegraphics[width=\linewidth]{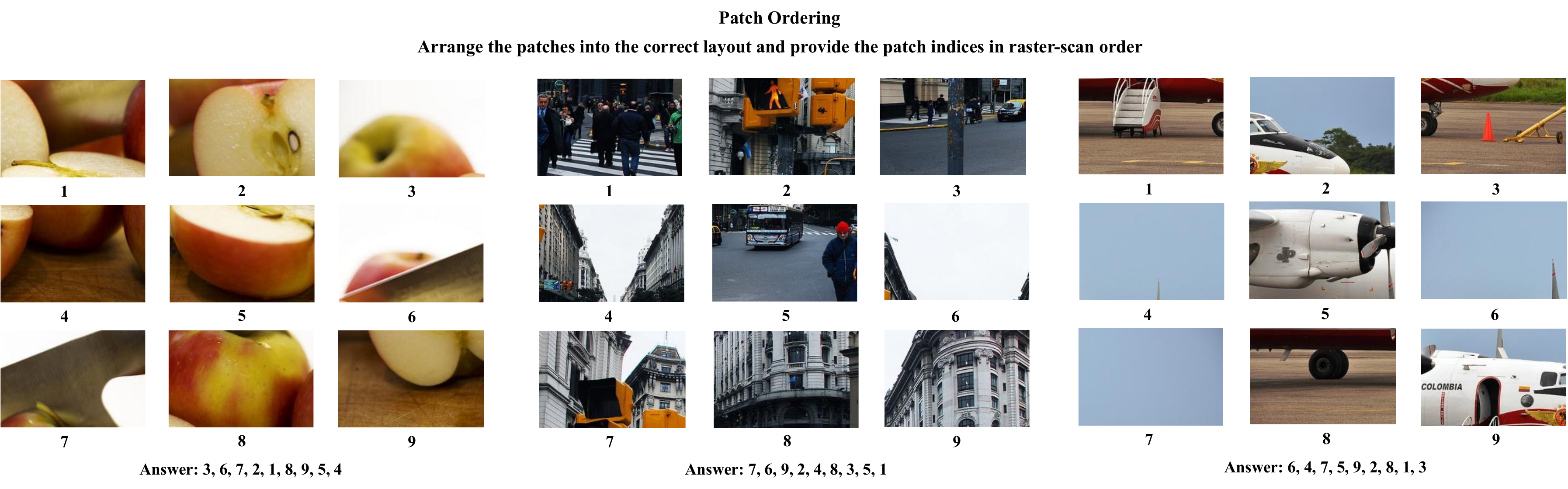}
    \caption{\textbf{Examples of the Patch Ordering task.} The ground-truth answer for each example is provided at the bottom.}
    \label{fig:task_ordering}
\end{figure*}

\begin{figure*}[ht]
    \centering
    \includegraphics[width=\linewidth]{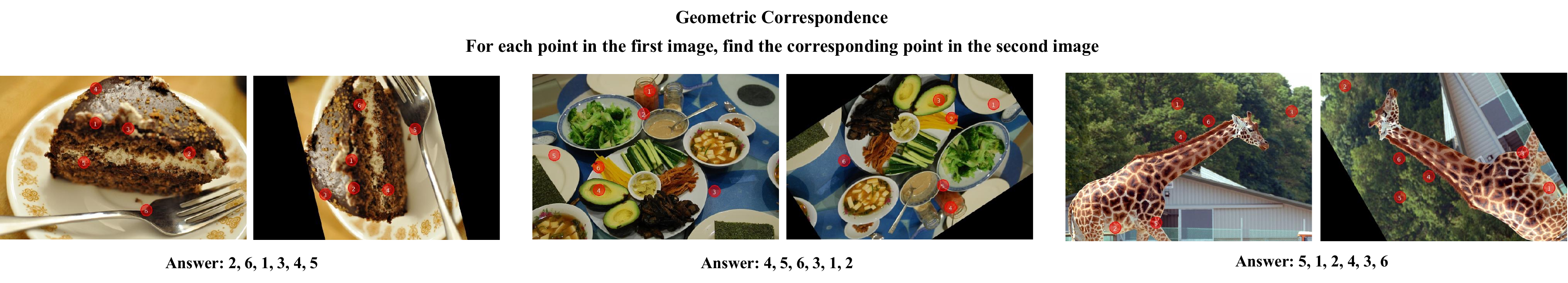}
    \caption{\textbf{Examples of the Geometric Correspondence task.} The ground-truth answer for each example is provided at the bottom.}
    \label{fig:task_correspondence}
\end{figure*}

We provide examples for each self-supervised task in Fig.~\ref{fig:task_rotation}, Fig.~\ref{fig:task_similarity}, Fig.~\ref{fig:task_inpainting}, Fig.~\ref{fig:task_ordering}, and Fig.~\ref{fig:task_correspondence}, respectively.

\section{Qualitative Results}
\label{sec:qualitative_results}

\begin{figure*}[ht]
    \centering
    \includegraphics[width=\linewidth]{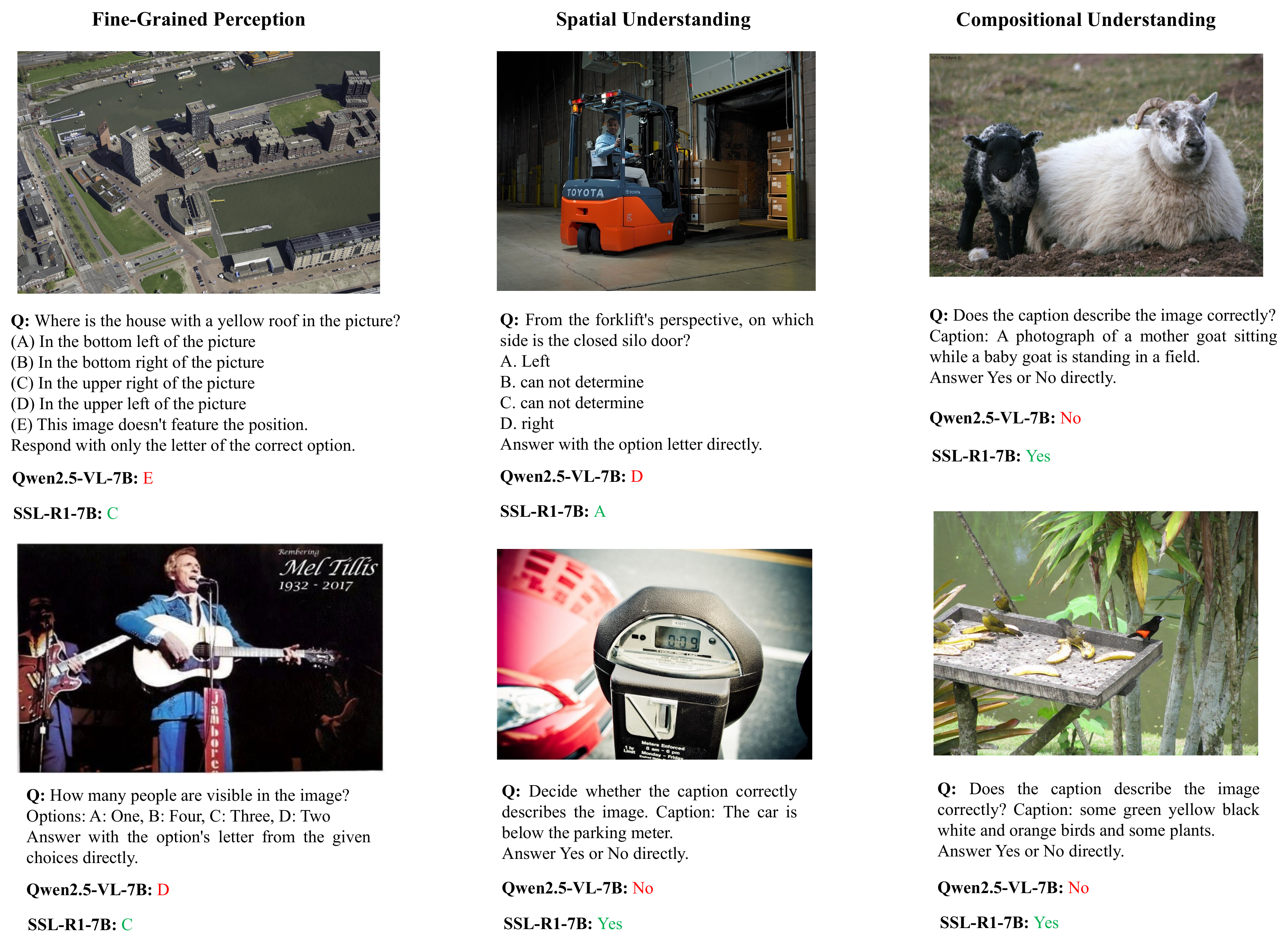}
    \caption{\textbf{Qualitative examples on three types of vision-centric multimodal benchmarks.} The wrong answers are marked in {\color{Red}red} while the correct answers are marked in {\color{Green}green}.}
    \label{fig:vlz}
\end{figure*}

We provide some qualitative examples of the baseline model vs. our SSL-R1 on vision-centric multimodal benchmarks in Fig.~\ref{fig:vlz}.

\end{document}